\documentclass[runningheads]{llncs}
\pdfoutput=1

\usepackage[mathletters]{ucs}
\usepackage[utf8x]{inputenc}
\usepackage[english]{babel}
\usepackage{lipsum}
 
\usepackage{microtype}
\usepackage{tikz}
\usetikzlibrary{positioning, fit, calc, decorations.pathmorphing}

\usepackage[disable]{todonotes}

\usepackage[cache=false,draft]{minted}
\usepackage{subfiles}
\usepackage[hidelinks]{hyperref}
\usepackage{cleveref}

\setminted{fontsize=\footnotesize}

\title{The Tactician (extended version)\thanks{This work was supported by the European Regional
    Development Fund under the project AI\&Reasoning (reg. no.
    CZ.02.1.01/0.0/0.0/15\_003/0000466) and by the \textit{AI4REASON} ERC
    Consolidator grant nr. 649043. }} \subtitle{A Seamless, Interactive Tactic
  Learner and Prover for Coq}

\author{Lasse Blaauwbroek\inst{1, 2} \and Josef Urban\inst{1} \and Herman
  Geuvers\inst{2}}
\authorrunning{L. Blaauwbroek et al.}
\institute{Czech Technical University, Prague, Czech Republic
  \and Radboud University, Nijmegen, The Netherlands \\
  \vspace{3pt} \email{lasse@blaauwbroek.eu, josef.urban@gmail.com,
    herman@cs.ru.nl}}

\makeatletter
\hypersetup{
  pdftitle={The Tactician (extended version): A Seamless, Interactive Tactic
  Learner and Prover for Coq},
  pdfsubject={Machine Learning for Theorem Provers},
  pdfauthor={Lasse Blaauwbroek, Josef Urban and Herman Geuvers},
  pdfkeywords={Coq Proof Assistant, Interactive Theorem Proving,
    Machine Learning, Tactic Search, Proof Synthesis}}
\makeatother

\begin{document}
\maketitle 
\begin{abstract}
  We present Tactician, a tactic learner and prover for the Coq Proof Assistant.
  Tactician helps users make tactical proof decisions while they retain
  control over the general proof strategy. To this end, Tactician learns
  from previously written tactic scripts and gives users either suggestions
  about the next tactic to be executed or altogether takes over the burden of
  proof synthesis. Tactician's goal is to provide users with a seamless,
  interactive, and intuitive experience together with robust and adaptive proof
  automation. In this paper, we give an overview of Tactician from the user's
  point of view, regarding both day-to-day usage and issues of package
  dependency management while learning in the large. Finally, we give a peek
  into Tactician's implementation as a Coq plugin and machine
  learning platform.

\end{abstract}

\section{Introduction} The Coq Proof
Assistant~\cite{the_coq_development_team_2019} is an Interactive Theorem Prover
in which one proves lemmas using tactic scripts. Individual tactics in these
scripts represent actions that transform the proof state of the lemma currently
being proved. A wide range of tactics exist, with a wide range of
sophistication. Basic tactics such as \texttt{apply \textit{lem}} and
\texttt{rewrite \textit{lem}} use an existing lemma \textit{lem} to perform one
specific inference or rewriting step while tactics like \verb|ring| and
\verb|tauto| implement entire decision procedures that are guaranteed to succeed
within specific domains. Finally, open-ended search procedures are implemented
by tactics like \verb|auto| and \verb|firstorder|. They can be used in almost
every domain but usually only work on simple proof states or need to be
calibrated carefully. Users are also encouraged to define new tactics that
represent basic steps, decision procedures, or specialized search procedures
within their specific mathematical domain.

When proving a lemma, the user's challenge is to observe the current proof state
and select the appropriate tactic and its arguments to be used. Often the user
makes this decision based on experience with previous proofs. If the current
proof state is similar to a previously encountered situation, then one can
expect that an effective tactic in that situation might also be effective now.
Hence, the user is continuously matching patterns of proof states in their mind
and selects the correct tactic based on these matches.

That is not the only task the user performs, however. When working on a
mathematical development, the user generally has two roles: (1) As a {\em
  strategist}, the user comes up with appropriate lemmas and sometimes decides
on the main structure of complicated proofs. (2) As a {\em tactician}, the user
performs the long and somewhat mindless process of mental pattern matching on
proof states, applying corresponding tactics until the lemma is proved. Many of
the steps in the tactician's role will be considered as ``obvious'' by a
mathematician. Our system is meant to replicate the pattern matching process
performed in this role, alleviating the user from this burden. Hence, we have
aptly named it Tactician.

To perform its job, Tactician can learn from existing proofs, by looking at how
tactics modify the proof state. Then, when proving a new lemma, the user can ask
the system to recommend previously used tactics based on the current proof state
and even to complete the whole proof using a search procedure based on these
tactic recommendations.

In our previous publication, the underlying machine learning and proof search
techniques employed by Tactician and how suitable data is extracted from Coq are
described~\cite{blaauwbroek2020tactic}. It also contains an evaluation of
Tactician's current proof automation performance on Coq's standard library. We
will not repeat these details here. Instead, we will focus on the operational
aspects and description of Tactician when used as a working and research tool.
\Cref{sec:system-overview} gives a mostly non-technical overview of the system
suitable for casual Coq users. That includes Tactician's design principles, its
mode of operation, a concrete example and a discussion on using Tactician in
large projects. \Cref{sec:technical-implementation} briefly discusses some of
Tactician's technical implementation issues, and
\Cref{sec:tactician-as-a-machine-learning-platform} describes how Tactician can
be used as a machine learning platform. Finally, \Cref{sec:related-work}
compares Tactician to related work. Installation instructions of Tactician can
be found at the project's website \url{http://coq-tactician.github.io}. This
pre-print is an extended version of our CICM paper with the same
title~\cite{DBLP:conf/mkm/BlaauwbroekUG20}.

\section{System Overview}
\label{sec:system-overview}
In this section, we give a mostly non-technical overview of Tactician suitable
for casual Coq users. \Cref{sec:design-decisions} states the guiding design
principles of the project, and \Cref{sec:mode-of-operation} describes the
resulting user workflow. On the practical side, \Cref{sec:a-concrete-example}
gives a simple, concrete example of Tactician's usage, while
\Cref{sec:learning-in-the-large} discusses how to employ Tactician in large
projects.

\subsection{Design Principles}
\label{sec:design-decisions}

For our system, we start with the principal goal of learning from previous
proofs to aid the user with proving new lemmas. In Coq, there are essentially
two notions of proof: (1) proof terms expressed in the Gallina language (Coq's
version of the Calculus of Inductive
Constructions~\cite{DBLP:conf/tlca/Paulin-Mohring93}); (2) tactic proof scripts
written by the user that can then generate a Gallina term. In principle, it is
possible to use machine learning on both notions of proof. We have chosen to
learn from tactic proof scripts for two reasons:
\begin{enumerate}
\item Tactics scripts are a higher-level and forgiving environment,
  which is more suitable for machine learning. A Gallina proof term must be
  generated extremely precisely while a tactic script often still works after
  minor local mutations have occurred. Gallina terms are also usually much
  bigger than their corresponding tactic script because individual tactics can
  represent large steps in a proof.
\item We acknowledge that automation systems within a proof assistant often
  still need input from the user to fully prove a lemma. Working on the tactic
  level allows the user to introduce domain-specific information to aid the
  system. For example, one can write new tactics that represent decision
  procedures and heuristics that solve problems Tactician could not otherwise
  solve. One can teach Tactician about such new tactics merely by using them in
  hand-written proofs a couple of times, after which the system will
  automatically start to use them.
\end{enumerate}

Apart from the principal goal described above, the most important objective of
Tactician is to be usable and remain usable by actual Coq users. Hence, we
prioritize the system's ``look and feel'' over its hard performance numbers. To
achieve this usability, Tactician needs to be pleasant to all parties involved,
which we express in four basic ``friendliness'' tenets.
 
\begin{description}
\item[User Friendly] If the system is to be used by actual Coq users, it should
  function as seamlessly as possible. After installation, it should be ready to
  go with minimal configuration and without needing to spend countless hours
  training a sophisticated machine learning model. Instead, there should be a
  model that can learn on the fly, with a future possibility to add a more
  sophisticated model that can be trained in batch once a development has been
  finished. Finally, all interaction with the system should happen within the
  standard Coq environment, regardless of which editor is used and without the
  need to execute custom scripts.
\item[Installation Friendly] Ease of installation is essential to reach solid
  user adoption. To facilitate this, the system should be implemented in Ocaml
  (Coq's implementation language), with no dependencies on machine learning
  toolkits written in other languages like Python or Matlab. Compilation and
  installation will then be just as easy as with a regular Coq release.
\item[Integration Friendly] The system should not be a fork of the main Coq
  codebase that has to be maintained separately. A fork would deter users from
  installing it and risk falling behind the upstream code. Instead, it should
  function as a plugin that can be compiled separately and then loaded into Coq
  by the user.
\item[Maintenance Friendly] We intend for Tactician not to become abandonware
  after main development has ceased, and at least remain compatible with the
  newest release of Coq. As a first step, the plugin should be entered into the
  Coq Package Index~\cite{coq_package_index}, enabling continuous integration
  with future versions of Coq. Additionally, assuming that Tactician becomes
  wildly popular, we eventually intend for it to be absorbed into the main Coq
  codebase.
\end{description}

\subsection{Mode of Operation}
\label{sec:mode-of-operation}

Analogously to Coq's mode of operation, Tactician can function both in
interactive mode and in compilation mode.

\subsubsection{Interactive Mode}

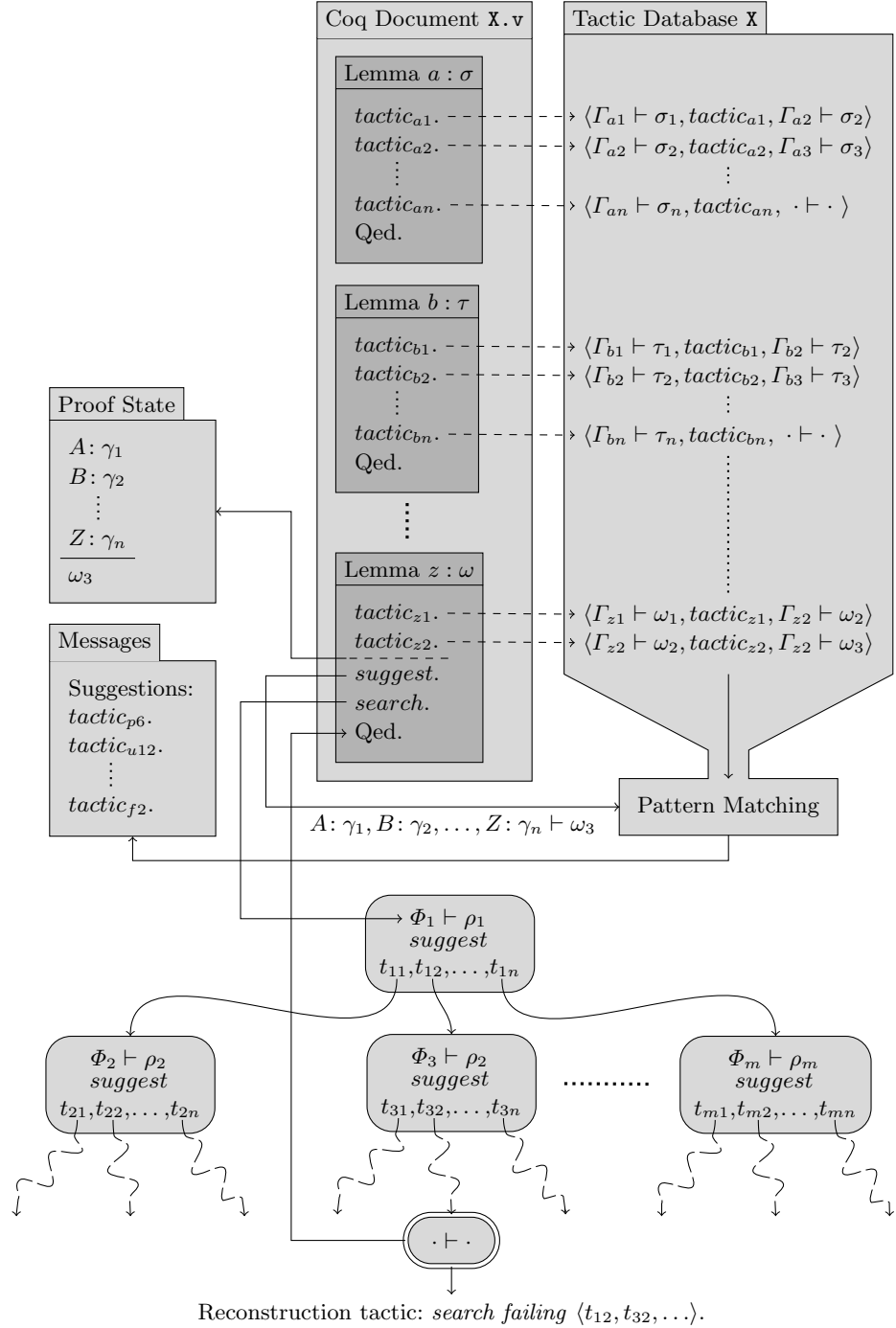
\begin{figure}
  \centering
  \begin{tikzpicture}
    \pgfdeclarelayer{bg1}
    \pgfdeclarelayer{bg2}
    \pgfsetlayers{bg1,bg2,main}

    \node [] (ta1) {$tactic_{a1}.$};
    \node [below = 2pt of ta1.base west, anchor = north west] (ta2) {$tactic_{a2}.$};
    \node [below = 10pt of ta2.south west, anchor = north west] (ta3) {$tactic_{an}.$};
    \node [below = 2pt of ta3.base west, anchor = north west] (qed1) {Qed.};
    \draw [dotted, thick] ([xshift = 20pt]ta2.south west) -- ([xshift = 20pt]ta3.north west);
    \node [fit=(ta1)(ta2)(ta3)(qed1), inner sep = 4pt] (lemmaa-box1) {};
    \node [above = 0pt of lemmaa-box1.north west, anchor = south west, fill = black!30] (lemmaa) {Lemma $a : \sigma$};
    \draw (lemmaa.south west) -- (lemmaa.north west) -- (lemmaa.north east) -- (lemmaa.south east);
    \begin{pgfonlayer}{bg2}
      \node [fit={(lemmaa-box1) (lemmaa.south east)}, inner sep = 0pt, fill = black!30, draw] (lemmaa-box1) {};
    \end{pgfonlayer}
    
    \node [below = 30 pt of qed1.south west, anchor = north west] (tb1) {$tactic_{b1}.$};
    \node [below = 2pt of tb1.base west, anchor = north west] (tb2) {$tactic_{b2}.$};
    \node [below = 10pt of tb2.south west, anchor = north west] (tb3) {$tactic_{bn}.$};
    \node [below = 2pt of tb3.base west, anchor = north west] (qed2) {Qed.};
    \draw [dotted, thick] ([xshift = 20pt]tb2.south west) -- ([xshift = 20pt]tb3.north west);
    \node [fit=(tb1)(tb2)(tb3)(qed2), inner sep = 4pt] (lemmab-box1) {};
    \node [above = 0pt of lemmab-box1.north west, anchor = south west, fill = black!30] (lemmab) {Lemma $b : \tau$};
    \draw (lemmab.south west) -- (lemmab.north west) -- (lemmab.north east) -- (lemmab.south east);
    \begin{pgfonlayer}{bg2}
      \node [fit={(lemmab-box1) (lemmab.south east)}, inner sep = 0pt, fill = black!30, draw] (lemmab-box2) {};
    \end{pgfonlayer}
    
    \node [below = 45 pt of qed2.south west, anchor = north west] (tz1) {$tactic_{z1}.$};
    \node [below = 2pt of tz1.base west, anchor = north west] (tz2) {$tactic_{z2}.$};
    \node [below = 0pt of tz2.south west, anchor = north west] (tp) {$suggest.$};
    \node [below = 2pt of tp.base west, anchor = north west] (ts) {$search.$};
    \node [below = 2pt of ts.base west, anchor = north west] (qed3) {Qed.};
    \node [fit=(tz1)(qed3), inner sep = 4pt] (lemmaz-box1) {};
    \node [above = 0pt of lemmaz-box1.north west, anchor = south west, fill = black!30] (lemmaz) {Lemma $z : \omega$};
    \draw (lemmaz.south west) -- (lemmaz.north west) -- (lemmaz.north east) -- (lemmaz.south east);
    \begin{pgfonlayer}{bg2}
      \node [fit={(lemmaz-box1) (lemmaz.south east)}, inner sep = 0pt, fill = black!30, draw] (lemmaz-box2) {};
    \end{pgfonlayer}

    \draw [dotted, very thick, shorten >= 4pt, shorten <= 4pt] (lemmab-box2.south) -- (lemmab-box2 |- lemmaz.north);
    
    \node [fit=(lemmaa) (lemmaz-box2), inner sep = 7pt] (documentx-box1) {};
    \node [above = 0pt of documentx-box1.north west, anchor = south west, fill = black!15] (documentx) {Coq Document \texttt{X.v}};
    \draw (documentx.south west) -- (documentx.north west) -- (documentx.north east) -- (documentx.south east);
    \begin{pgfonlayer}{bg1}
      \node [fit={(documentx-box1) (documentx.south east)}, fill = black!15, draw, inner sep = 0pt] (documentx-box2) {};
    \end{pgfonlayer}

    \node [right = 50pt of ta1] (ta1db)
    {$\langle \Gamma_{a1} \vdash \sigma_1, tactic_{a1}, \Gamma_{a2} \vdash \sigma_2 \rangle$};
    \path (ta2.north east) -| node [anchor = north west] (ta2db)
    {$\langle \Gamma_{a2} \vdash \sigma_2, tactic_{a2}, \Gamma_{a3} \vdash \sigma_3 \rangle$} (ta1db.west);
    \path (ta3.north east) -| node [anchor = north west] (ta3db)
    {$\langle \Gamma_{an} \vdash \sigma_n, tactic_{an}, \,\,\cdot \vdash \cdot\,\, \rangle$} (ta1db.west);
    \path (tb1.north east) -| node [anchor = north west] (tb1db)
    {$\langle \Gamma_{b1} \vdash \tau_1, tactic_{b1}, \Gamma_{b2} \vdash \tau_2 \rangle$} (ta1db.west);
    \path (tb2.north east) -| node [anchor = north west] (tb2db)
    {$\langle \Gamma_{b2} \vdash \tau_2, tactic_{b2}, \Gamma_{b3} \vdash \tau_3 \rangle$} (ta1db.west);
    \path (tb3.north east) -| node [anchor = north west] (tb3db)
    {$\langle \Gamma_{bn} \vdash \tau_n, tactic_{bn}, \,\,\cdot \vdash \cdot\,\, \rangle$} (ta1db.west);
    \path (tz1.north east) -| node [anchor = north west] (tz1db)
    {$\langle \Gamma_{z1} \vdash \omega_1, tactic_{z1}, \Gamma_{z2} \vdash \omega_2 \rangle$} (ta1db.west);
    \path (tz2.north east) -| node [anchor = north west] (tz2db)
    {$\langle \Gamma_{z2} \vdash \omega_2, tactic_{z2}, \Gamma_{z2} \vdash \omega_3 \rangle$} (ta1db.west);
    \draw [dashed, ->]
    (ta1) -- (ta1db);
    \draw [dashed, ->](ta2) -- (ta2db);
    \draw [dashed, ->](ta3) -- (ta3db);
    \draw [dashed, ->](tb1) -- (tb1db);
    \draw [dashed, ->](tb2) -- (tb2db);
    \draw [dashed, ->](tb3) -- (tb3db);
    \draw [dashed, ->](tz1) -- (tz1db);
    \draw [dashed, ->](tz2) -- (tz2db);
    \draw [dotted, thick]
    (ta2db.south) -- (ta2db.south |- ta3db.north)
    (ta2db.south |- tb2db.south) -- (ta2db.south |- tb3db.north)
    (ta2db.south |- tb3db.south) -- (ta2db.south |- tz1db.north);

    \node [fit=(ta1db) (ta2db) (ta3db) (tz2db), inner sep = 4pt] (database-box) {};
    \path (documentx.north east) -| node [anchor = north west, fill=black!15] (database) {Tactic Database \texttt{X}} (database-box.north west);
    \draw (database.south west) |- (database.north east) -- (database.south east);
    \node [below = 10pt of database-box |- documentx-box2.south, anchor = center, inner sep = 7pt] (pt) {Pattern Matching};
    \draw [->] (database-box.south) -- (pt);
    \begin{pgfonlayer}{bg1}
      \draw[fill = black!15]
      (database-box.south west) -- (database.south west) -| (database-box.south east) --
      ($(database-box.south)+(8pt, -30pt)$) |- (pt.north east) |- (pt.south west) -- (pt.north west) -|
      ($(database-box.south)+(-8pt, -30pt)$) -- (database-box.south west);
    \end{pgfonlayer}

    \draw[->] (tp.west) -- ($(tp.west -| documentx-box2.west)+(-20pt,0)$) |- node[near end, below, xshift=4pt] {$A\!:\gamma_1,B\!:\gamma_2,\ldots,Z\!:\gamma_n\vdash\omega_3$} (pt);

    \node [left = 44pt of pt -| documentx-box2.west, anchor = east, text width = 50.6pt] (mtn) {$tactic_{f2}.$};
    \node [above = 10pt of mtn.north west, anchor = south west] (mt2) {$tactic_{u12}.$};
    \node [above = 2pt of mt2.north west, anchor = base west] (mt1) {$tactic_{p6}.$};
    \node [above = 2pt of mt1.north west, anchor = base west] (mpd) {Suggestions:};
    \draw [dotted, thick] ([xshift = 20pt]mt2.south west) -- ([xshift = 20pt]mtn.north west);
    \node [fit=(mt1)(mt2)(mtn)(mpd), inner sep = 4pt] (messages-box1) {};
    \node [above = 0pt of messages-box1.north west, anchor = south west, fill = black!15] (messages) {Messages};
    \draw (messages.south west) -- (messages.north west) -| (messages.south east);
    \begin{pgfonlayer}{bg2}
      \node [fit={(messages-box1) (messages.south east)}, inner sep = 0pt, fill = black!15, draw] (messages-box2) {};
    \end{pgfonlayer}

    \draw[->] (pt.south) -- ($(pt.south)+(0, -10pt)$) -| (messages-box2.south);

    \node [above = 45pt of mpd.base west, anchor = base west] (psc) {$\omega_3$};
    \node [above = 2pt of psc.north west, anchor = south west] (pshn) {$Z\!:\gamma_n$};
    \node [above = 10pt of pshn.north west, anchor = south west] (psh2) {$B\!:\gamma_2$};
    \node [above = 2pt of psh2.north west, anchor = base west] (psh1) {$A\!:\gamma_1$};
    \draw [dotted, thick] ([xshift = 15pt]psh2.south west) -- ([xshift = 15pt]pshn.north west);
    \draw (pshn.south west) -- (pshn.south east);
    \node [fit=(psc)(pshn)(psh2)(psh1), inner sep = 4pt] (ps-box1) {};
    \node [above = 0pt of ps-box1.north west, anchor = south west, fill = black!15] (ps) {Proof State};
    \draw (ps.south west) -- (ps.north west) -| (ps.south east);
    \begin{pgfonlayer}{bg2}
      \node [fit={(ps-box1)(ps.south east)(messages-box1.east |- ps-box1)}, inner sep = 0pt, fill = black!15, draw] (ps-box2) {};
    \end{pgfonlayer}
    \draw[dashed] (tp.north east) -- (tp.north west);
    \draw[->] (tp.north west) -- ($(tp.north west -| documentx-box2.west)+(-10pt,0)$) |- (ps-box2.east);

    \node[fit={(documentx)(pt)(ps)(database-box)($(pt.south)+(0,-10pt)$)}, inner sep = 0] (all-documents) {};

    \tikzset{
      pics/searchnode/.style n args={4}{
        code={
          \begin{scope}
            \node (-predict) {$suggest$};
            \node[inner sep = 0, opacity = 0, below = 0pt of -predict] (-phantom) {#2,#3,\ldots,#4};
            \node[inner sep = 0, right = 0pt of -phantom.west] (-t1) {#2,};
            \node[inner sep = 0, right = 0pt of -t1.base east, anchor = base west] (-t2) {#3,};
            \node[inner sep = 0, right = 0pt of -t2.base east, anchor = base west] (-dots) {\ldots};
            \node[inner sep = 0, right = 0pt of -dots.base east, anchor = base west] (-tn) {,#4};
            \node[fit=(-t1)(-t1)(-tn)] (-ts) {};
            \node[above = 0pt of -predict.north, anchor = base] (-state) {#1};
            \begin{pgfonlayer}{bg1}
              \node[rounded corners = 10pt, fit=(-ts)(-predict)(-state), inner sep = 2pt, draw, fill=black!15] (-boundary) {};
            \end{pgfonlayer}
          \end{scope}
        }},
      unbounded/.style = {
        ->, dash pattern = on 10pt off 2pt,
        decorate, decoration={
          snake, segment length=16pt, amplitude = 5pt,
          pre=curveto, pre length = 6pt, post=curveto, post length = 1pt}}}
    
    \pic[below = 25pt of all-documents, xshift=-9pt] (s1) {searchnode={$\Phi_{1}\vdash\rho_1$}{$t_{11}$}{$t_{12}$}{$t_{1n}$}{}};
    \pic[below = 28pt of s1-boundary] (s3) {searchnode={$\Phi_{3}\vdash\rho_2$}{$t_{31}$}{$t_{32}$}{$t_{3n}$}{}};
    \pic[left = 75pt of s3-boundary] (s2) {searchnode={$\Phi_{2}\vdash\rho_2$}{$t_{21}$}{$t_{22}$}{$t_{2n}$}{}};
    \pic[right = 75pt of s3-boundary] (sm) {searchnode={$\Phi_{m}\vdash\rho_m$}{$t_{m1}$}{$t_{m2}$}{$t_{mn}$}{}};
    \coordinate[below = 32.5pt of s2-boundary] (s5);
    \coordinate[left = 45pt of s5] (s4);
    \coordinate[right= 45pt of s5] (s6);
    \node[below = 32.5pt of s3-boundary, rounded corners = 10pt, draw, double, double distance = 1.5pt, fill=black!15, inner sep = 7pt] (s8) {$\,\,\cdot\vdash \cdot\,\,$};
    \coordinate[left = 45pt of s8.north] (s7);
    \coordinate[right= 45pt of s8.north] (s9);
    \coordinate[below = 32.5pt of sm-boundary] (s11);
    \coordinate[left = 45pt of s11] (s10);
    \coordinate[right= 45pt of s11] (s12);
    \draw[->] (ts.west) -- ($(ts.west -| documentx-box2.west)+(-30pt,0)$) |- (s1-state);
    \draw[<-] (qed3.west) -- ($(qed3.west -| documentx-box2.west)+(-10pt,0)$) |- (s8);
    \draw[->] (s1-t1.south) to[in=90, out=270] (s2-boundary.north);
    \draw[->] (s1-t2.south) to[in=90, out=270] (s3-boundary.north);
    \draw[->] (s1-tn.south) to[in=90, out=270] (sm-boundary.north);
    \draw[unbounded] (s2-t1.south) to[in=90, out=270] (s4);
    \draw[unbounded] (s2-t2.south) to[in=90, out=270] (s5);
    \draw[unbounded] (s2-tn.south) to[in=90, out=270] (s6);
    \draw[unbounded] (s3-t1.south) to[in=90, out=270] (s7);
    \draw[unbounded] (s3-t2.south) to[in=90, out=270] (s8);
    \draw[unbounded] (s3-tn.south) to[in=90, out=270] (s9);
    \draw[unbounded] (sm-t1.south) to[in=90, out=270] (s10);
    \draw[unbounded] (sm-t2.south) to[in=90, out=270] (s11);
    \draw[unbounded] (sm-tn.south) to[in=90, out=270] (s12);
    \draw[dotted, very thick, shorten >=11pt, shorten <= 11pt] (s3-boundary) -- (sm-boundary);

    \node[below=11pt of s8] (reconstruct) {Reconstruction tactic: \textit{search failing} $\langle t_{12},t_{32},\ldots\rangle$.};
    \draw[->] (s8) -- (reconstruct);

  \end{tikzpicture} 
  \caption{A schematic overview of Tactician in its interactive mode of operation.}
  \label{fig:tactician-interactive-overview}
\end{figure}

We illustrate the interactive mode of operation of Tactician using the schematic
in \Cref{fig:tactician-interactive-overview}. When the user starts a new Coq
development file---say \texttt{X.v}---the first thing Tactician does is create
an (in-memory) empty tactic database \texttt{X} corresponding to this file. The
user then starts to prove lemmas as usual. Behind the scenes, every executed
tactic, e.g. $tactic_{a1}$, is saved into the database accompanied by the proof
states before and after tactic execution, in this case,
$\langle\Gamma_{a1}\vdash\sigma_1, tactic_{a1}, \Gamma_{a2}\vdash\sigma_2
\rangle$. The difference between these two states represents the action
performed by the tactic, while the state before the tactic represents the
context in which it was useful. By recording many such triples for a tactic, we
create a dataset representing an approximation of that tactic's semantic
meaning. The database is kept synchronized with the user's movement within the
document throughout the entire interactive session.

After proving a few lemmas by hand, the user can start to reap the fruits of the
database. For this, the tactics \texttt{suggest} and \texttt{search} are
available. We illustrate their use in the schematic when ``Lemma z : $\omega$''
is being proven. The user first executes two normal tactics. After that, Coq's
proof state window displays a state for which the user is unsure what tactic to
use. Here Tactician's tactics come in.
\begin{description}
\item[\texttt{suggest}] This tactic can be executed to ask Tactician for a list
  of recommendations. The current proof state
  $A\!:\gamma_1,B\!:\gamma_2,\ldots,Z\!:\gamma_n\vdash \omega_3$ is fed into the
  pattern matching engine, which will perform a comparison with the states in
  the tactic database. From this, an ordered list of recommendations is
  generated and displayed in Coq's messages window, where the user can select a
  tactic to execute.
\item[\texttt{search}] Alternatively, the system can be asked to \texttt{search}
  for a complete proof. We start with the current proof state, which we rename
  to $\Phi_1 \vdash \rho_1$ for clarity. Then a search tree is formed by
  repeatedly running \texttt{suggest} on the proof state and executing the
  suggested tactics. This tree can be traversed in various ways, finishing only
  when a complete proof has been found.

  If a proof is found, two things happen. (1) The Gallina proof term that is
  found is immediately submitted to Coq's proof engine, after which the proof
  can be closed with \texttt{Qed}. (2) Tactician generates a reconstruction
  tactic \texttt{search failing $\langle \texttt{t}_{12}, \texttt{t}_{32},...
    \rangle$} which is displayed to the user (see the bottom of the figure). The
  purpose of this tactic is to provide a modification resilient proof cache that
  also functions when Tactician is not installed. Its semantics is to first try
  to use the previously found list of tactics \texttt{$\langle \texttt{t}_{12},
    \texttt{t}_{32},... \rangle$} to complete the proof immediately. ``Failing''
  that (presumably due to changes in definitions or lemmas), a new search is
  initiated to recover the proof. To use the cache, the user should copy it and
  replace the original \texttt{search} invocation with it in the source file.
\end{description}

\subsubsection{Compilation Mode}

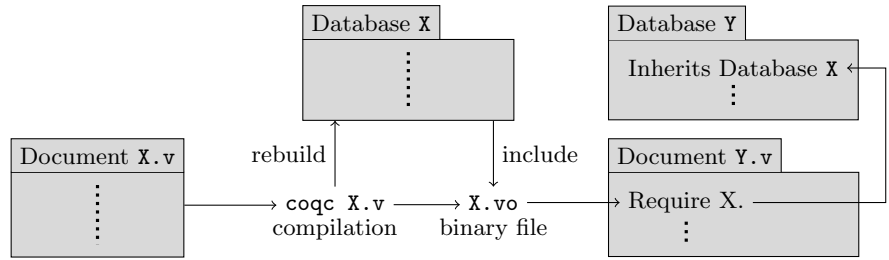
\begin{figure}
  \centering
  \begin{tikzpicture}
    \pgfdeclarelayer{bg1}
    \pgfdeclarelayer{bg2}
    \pgfsetlayers{bg1,bg2,main}
    
    \node [] (inherit) {Inherits Database \texttt{X}};
    \coordinate [below = 9pt of inherit] (dbydot);
    \draw [dotted, very thick] (inherit) -- (dbydot);
    \node [fit=(inherit)(dbydot), inner sep = 4pt] (databasey-box1) {};
    \node [above = 0pt of databasey-box1.north west, anchor = south west, fill = black!15] (databasey) {Database \texttt{Y}};
    \draw (databasey.south west) -- (databasey.north west) -| (databasey.south east);
    \begin{pgfonlayer}{bg2}
      \node [fit={(databasey-box1) (databasey.south east)}, inner sep = 0pt, fill = black!15, draw] (databasey-box2) {};
    \end{pgfonlayer}
    
    \node [below = 50pt of inherit.base west, anchor = base west] (require) {Require X.};
    \coordinate [below = 9pt of require] (docydot);
    \draw [dotted, very thick] (require) -- (docydot);
    \node [fit=(require)(docydot), inner sep = 4pt] (documenty-box1) {};
    \node [above = 0pt of documenty-box1.north west, anchor = south west, fill = black!15] (documenty) {Document \texttt{Y.v}};
    \draw (documenty.south west) -- (documenty.north west) -| (documenty.south east);
    \begin{pgfonlayer}{bg2}
      \node [fit={(documenty-box1)(documenty.south east)(documenty-box1.east -| databasey-box2.east)}, inner sep = 0pt, fill = black!15, draw] (documenty-box2) {};
    \end{pgfonlayer}
    
    \node [left = 35pt of require.west, label={base:binary file}] (xvo) {\verb|X.vo|};
    \node [left = 25pt of xvo.base west, anchor = base east, label={base:compilation}] (coqc) {\texttt{coqc X.v}};

    \coordinate [left = 35pt of coqc.west |- documenty-box2.south west] (documentx-box1);
    \node [above = 0pt of documentx-box1 |- documenty.south west, anchor = south east, fill = black!15] (documentx) {Document \texttt{X.v}};
    \draw (documentx.south west) |- (documentx.north east) -- (documentx.south east);
    \begin{pgfonlayer}{bg1}
      \node [fit={(documentx-box1) (documentx.south west)}, fill = black!15, draw, inner sep = 0pt] (documentx-box2) {};
    \end{pgfonlayer}
    \draw [dotted, very thick, shorten >= 5pt, shorten <= 5pt] (documentx-box2.north) -- (documentx-box2.south);

    \coordinate [left = 5pt of xvo.east |- databasey-box2.south west] (database-box) {};
    \node [right = 10pt of coqc.west |- databasey-box2.north west, anchor = south west, fill=black!15] (database) {Database \texttt{X}} (database-box.north west);
    \draw (database.south west) |- (database.north east) -- (database.south east);
    \begin{pgfonlayer}{bg1}
      \node [fit={(database-box)(database.south west)}, fill = black!15, draw, inner sep = 0pt] (database-box2) {};
    \end{pgfonlayer}
    \draw [dotted, very thick, shorten >= 5pt, shorten <= 5pt] (database-box2.north) -- (database-box2.south);

    \draw[<-] (coqc) -- (documentx-box2.east |- coqc.west);
    \draw[->] (coqc.mid east) -- (xvo.mid west);
    \draw[<-] (require.west) -- ($(require.west -| documenty-box2.west)+(-10pt,0pt)$) |- (xvo);
    \draw[->] (require.east) -- ($(require.east -| documenty-box2.east)+(10pt,0)$) |- (inherit);
    \draw[->] (coqc.north) -- node[left] {rebuild} (coqc.north |- database-box2.south);
    \draw[<-] (xvo.north) -- node[right] {include} (xvo.north |- database-box2.south);

  \end{tikzpicture} 
  \caption{A schematic overview of Tactician in its compilation mode of operation.}
  \label{fig:tactician-batch-overview}
\end{figure}

This mode is visualized in \Cref{fig:tactician-batch-overview}. After the file
\texttt{X.v} has been finished, one might want to depend on it in other files.
This requires the file to be compiled into a binary \texttt{X.vo} file. The
compilation is performed using the command \texttt{coqc X.v}. Tactician is
integrated into this process. During compilation, the tactic database is rebuilt
in the same way as in interactive mode and then included in the \texttt{.vo}
file. When development \texttt{X.v} is then \texttt{Require}d by another
development file \texttt{Y.v}, the tactic database of \texttt{X.v} is
automatically inherited.

\subsection{A Concrete Example}
\label{sec:a-concrete-example}
We now give a simple example use-case based on lists. Starting with an empty
file, Tactician is immediately ready for action. We proceed as usual by giving a
standard inductive definition of lists of numbers with their corresponding
notation and a function for concatenation.
\begin{minted}{coq}
Inductive list :=
| nil  : list
| cons : nat -> list -> list.
Notation "[]"      := nil.
Notation "x :: ls" := (cons x ls).

Fixpoint concat ls₁ ls₂ :=
match ls₁ with
| []      => ls₂
| x::ls₁' => x::(ls₁' ++ ls₂)
end where "ls₁ ++ ls₂" := (concat ls₁ ls₂).
\end{minted}
We wish to prove some standard properties of concatenation. The first is a lemma
stating that the empty list \texttt{[]} is the right identity of concatenation
(the left identity is trivial).
\begin{minted}{coq}
Lemma concat_nil_r : ∀ ls, ls ++ [] = ls.
\end{minted}
With Tactician installed, we immediately have access to the new tactics
\texttt{suggest} and \texttt{search}. Neither tactic will produce a result when
used now since the system has not had a chance to learn from proofs yet.
Therefore, we will have to prove this lemma by hand.

\begin{minted}{coq}
Proof.
intros. induction ls.
- simpl. reflexivity.
- simpl. f_equal. apply IHls.
Qed.
\end{minted}
The system has immediately learned from this proof (it was even learning during
the proof) and is now ready to help us with a proof of the associativity of
concatenation.
\begin{minted}{coq}
Lemma concat_assoc :
  ∀ ls₁ ls₂ ls₃, (ls ++ ls₂) ++ ls₃ = ls ++ (ls₂ ++ ls₃).
\end{minted}
Now, if we execute \texttt{suggest}, it outputs the ordered list \texttt{intros,
  simpl, f\_equal, reflexivity}. Indeed, using \texttt{intros} as our next
tactic is not unreasonable. We can repeatedly ask \texttt{suggest} for a
recommendation after every tactic we input, which sometimes gives us good
tactics and sometimes bad tactics. However, we can also eliminate the middle-man
and execute the \texttt{search} tactic, which immediately finds a proof.
\begin{minted}{coq}
Proof. search. Qed.
\end{minted}
To cache the proof that is found for the future, we can copy-paste the
reconstruction tactic that Tactician prints into the source file. This example
shows how the system can quickly learn from very little data and with minimal
effort from the user. Of course, this also scales to much bigger developments.

We continue our example for more advanced Coq users to showcase how Tactician
can learn to use custom domain-specific tactics. We begin by defining an
inductive property encoding that one list is a non-contiguous sublist of
another.
\begin{minted}{coq}
Inductive sublist : list -> list -> Prop :=
| sl_nil : sublist [] []
| sl_cons₁ ls₁ ls₂ n : sublist ls₁ ls₂ -> sublist ls₁ (n::ls₂)
| sl_cons₂ ls₁ ls₂ n : sublist ls₁ ls₂ -> sublist (n::ls₁) (n::ls₂).
\end{minted}
We now wish to prove that some lists have the sublist property. For example,
\texttt{sublist (9::3::[]) (4::7::9::3::[])}. Deciding this is not entirely
trivial, because it is not possible to judge from the head of the list whether
to apply \texttt{sl\_cons₁} or \texttt{sl\_cons₂}. Instead of manually writing
these proofs, we create a domain-specific, heuristic proving tactic that
automatically tries to find a proof.
\begin{minted}{coq}
Ltac solve_sublist := solve [match goal with
| |- sublist [] [] => apply sl_nil
| |- sublist (_::_) [] => fail
| |- sublist _ _ =>
       (apply sl_cons₁ + apply sl_cons₂); solve_sublist
| |- _ => solve [auto]
end].
\end{minted}
This tactic looks at the current proof state and checks that the goal is of the
form \texttt{sublist ls₁ ls₂}. If the lists are empty, it can be finished using
\texttt{sl\_nil}. If \texttt{ls₂} is empty but \texttt{ls₁} nonempty, the
postfix is unprovable, and the tactic fails. In all other cases, we initiate a
backtracking search where we try to apply either \texttt{sl\_cons₁} or
\texttt{sl\_cons₂} and recurse. Finally, we add a simple catch-all clause that
tries to prove any side-conditions using Coq's built-in \texttt{auto}. We now
teach Tactician about the existence of our previous lemmas and the
domain-specific tactic by defining some simple examples. Finally, we ask
Tactician to solve a more complicated, compound problem.
\begin{minted}{coq}
Lemma ex1 : sublist (9::3::[]) (4::7::9::3::[]).
Proof. solve_sublist. Qed.

Lemma ex2 : ∀ ls, 1::2::ls ++ [] = 1::2::ls.
Proof. intro. rewrite concat_nil_r. reflexivity. Qed.

Lemma dec2 : ∀ ls₁ ls₂, sublist ls₁ ls₂ ->
  sublist (7::9::13::ls₁) (8::5::7::[] ++ 9::13::ls₂ ++ []).
Proof. search. Qed.
\end{minted}
The proof found by Tactician is \texttt{rewrite
  concat\_nil\_r;intros;solve\_sublist}. It has automatically figured out that
it needs to introduce the hypothesis, normalize the list and then run the
domain-specific prover. This example is somewhat contrived but should illustrate
how the user can easily teach Tactician domain-specific knowledge.

\subsection{Learning and Proving in the Large}
\label{sec:learning-in-the-large}
The examples above are fun to play with and useful for demonstration purposes.
However, when using Tactician to develop real projects, three main issues need
to be taken care of, namely (1) instrumenting dependencies, (2) instrumenting
the standard library and, (3) reproducible builds. Below, we describe how to use
Tactician in complex projects and, in particular, how these issues are solved.

Tactician itself is a collection of easily installed Opam~\cite{opam} packages
distributed through the Coq Package Index~\cite{coq_package_index}. The package
\texttt{coq-tactician} provides the main functionality. It needs to be installed
to run the examples of \Cref{sec:a-concrete-example}. These examples have no
dependencies and make minimal use of Coq's standard library. All learning is
done within one file, making them a simple use-case. Things become more
complicated when one starts to use the standard library and libraries defined in
external dependencies. Although Tactician will keep working normally in those
situations, by default, it does not learn from proofs in these libraries. Hence,
Tactician's ability to synthesize proofs for lemmas concerning the domain of
those libraries will be severely limited.

The main question to remedy this situation is where Tactician should get its
learning data from. As explained in \Cref{sec:mode-of-operation}, Tactician
saves the tactic database of a library in the compiled \texttt{.vo} binaries.
This database becomes available as soon as the compiled library is loaded.
However, this only works if the library was compiled with Tactician enabled,
which is the case neither for Coq's standard library nor most external packages.
Hence, we need to instrument these libraries by recompiling them while Tactician
is loaded. Loading Tactician amounts to finding a way of convincing Coq to
pre-load the library \texttt{Tactician.Ltac1.Record} before starting the
compilation process.

\subsubsection{External Dependency Instrumentation}
An external dependency can be any collection of Coq source files together with
an arbitrary build system. Injecting the loading of
\texttt{Tactician.Ltac1.Record} into the build process automatically is not
possible in general. However, we do provide a simple solution for the most
common situation. Coq provides package developers with a utility called
\texttt{coq\_makefile} that automatically generates a Makefile to build and
install their files. This build system is usually packaged up with Opam to be
released on the Coq Package Index. Although this package index does not require
packages to use \texttt{coq\_makefile}, most do this in practice.

Makefiles generated by \texttt{coq\_makefile} are highly customizable through
environment variables. Tactician provides command-line utilities called
\texttt{tactician enable} and \texttt{tactician disable} that configure Opam to
automatically inject Tactician through these environment variables. When
building packages without Opam, the user can modify the environment by running
\texttt{eval \$(tactician inject)} before building. This solution will suffice
to instrument most packages created using \texttt{coq\_makefile}, as long as
authors do not customize the resulting build file too heavily. We will add
support for Coq's new Dune build system~\cite{dune} when it has stabilized. For
more stubborn packages, rather aggressive methods of injecting Tactician are
also possible but, in general, packages that circumvent instrumentation are
always possible. Therefore, we do not provide built-in solutions for those
cases.

\subsubsection{Standard Library Instrumentation}
In order to instrument Coq's standard library, it also needs to be recompiled
with \texttt{Tactician.Ltac1.Record} pre-loaded. We provide the Opam package
\texttt{coq-tactician-stdlib} for this purpose. This package does not contain
any code, but simply takes the source files of the installed standard library
and recompiles them. It then promptly commits the Cardinal Sin of Package
Management by overwriting the original binary \texttt{.vo} files of the standard
library. We defend this choice by noting that (1) the original files will be
backed up and restored when \texttt{coq-tactician-stdlib} is removed and (2) the
alternative of installing the recompiled standard library in a secondary
location is even worse. This choice would cause a rift in the users local
ecosystem of packages, with some packages relying on the original standard
library and some on the recompiled one. Coq will refuse to load two packages
from the rivaling ecosystems citing ``incompatible assumptions over the standard
library,'' forever setting them apart.

Even with our choice of overwriting the standard library, an ecosystem rift
still occurs if packages depending on Coq already pre-existed. To resolve this,
Tactician ships with the command-line utility \texttt{tactician recompile} that
helps the user find and recompile these packages.

\subsubsection{Tactician Usage within Packages}
In order to use Tactician's \texttt{suggest} and \texttt{search} tactics, the
library \texttt{Tactician.Ltac1.Tactics} needs to be loaded. However, we
strongly advise against loading this library directly, for two reasons. (1) If a
development \texttt{X} that uses Tactician is submitted to the Coq Package Index
as \texttt{coq-x}, an explicit dependency on \texttt{coq-tactician} is needed.
This dependency can be undesirable due to users potentially being unwilling to
install Tactician. (2) It would undermine the build reproducibility of the
package. Even though \texttt{coq-tactician} would be installed as a dependency
when \texttt{coq-x} is installed, there is no way to ensure that Tactician has
instrumented the other dependencies of the package. Hence, it is likely that
Tactician will be operating with a smaller tactic database, reducing its ability
to prove lemmas automatically.

Instead, the package \texttt{coq-x} should depend on
\texttt{coq-tactician-dummy}. This package is extremely simple, containing one
30-line library called \texttt{Tactician.\allowbreak Ltac1Dummy}. It provides
alternative versions of Tactician tactics that act as dummies of the real
version. Tactics \texttt{suggest} and \texttt{search} will not perform any
action. However, tactic \texttt{search failing $\langle$...$\rangle$}, described
in \Cref{sec:mode-of-operation}, will still be able to complete a proof using
its cache (but without the ability to search for new proofs in case of failure).
A released package can thus only employ cached searches. This way, any build
will be reproducible.

During development, the real version of Tactician should be loaded to gain its
full power. Instead of loading it explicitly through a \texttt{Require} in
source files, we recommend that users load it through the \texttt{coqrc} file.
Coq will automatically process any vernacular defined in this file at startup.
The command-line utility \texttt{tactician enable} will assist in adding the
correct vernacular to the \texttt{coqrc} file.

\section{Technical Implementation}
\label{sec:technical-implementation}

In this section, we provide a peek behind the curtains of Tactician's technical
implementation and how it is integrated with Coq. A previous publication already
covers the following aspects of Tactician~\cite{blaauwbroek2020tactic}: (1) The
machine learning models used to suggest tactics; (2) an explanation of how data
extracted from Coq is decomposed and transformed for these models; and (3) the
search procedure to synthesize new proofs. These details are therefore omitted
here.

\subsection{Intercepting Tactics}
\label{sec:intercepting-tactics}

Tactician is implemented as a plugin that provides a new proof mode (tactic
language) to Coq. This proof mode contains precisely the same syntactical elements
as the Ltac1 tactic language~\cite{DBLP:conf/lpar/Delahaye00}. The purpose of
the proof mode is to intercept and decompose executed tactics and save them in
Tactician's database. After interception, the tactics are redirected back to the
regular Ltac1 engine. By loading the library \texttt{Tactician.Ltac1.\allowbreak
  Record}, this proof mode is activated 
instead of the regular Ltac1 language.

Note that Ltac1 is the most popular but by no means the only tactic language
for
Coq~\cite{DBLP:journals/pacmpl/KaiserZKRD18,DBLP:conf/esop/MalechaB16,DBLP:journals/jfrea/GonthierM10,pedrot2019ltac2}.
All these languages are compiled into a proof monad implemented on the OCaml
level~\cite{DBLP:journals/jlp/KirchnerM10}. It would be preferable to instrument
the proof monad directly as this would enable us to record tactics from all
languages at once. It appears that this is impossible, though, because the
structure of the monadic interface does not allow us to recover high-level
tactical units such as decision procedures, even when implemented as the most
general Free Monad. As such, Tactician only supports Ltac1 at the moment. In the
future, we intent to provide improved support for
SSreflect~\cite{DBLP:journals/jfrea/GonthierM10} and support for recording the
new Ltac2 language~\cite{pedrot2019ltac2}.

\subsection{State Synchronization}
\label{sec:state-synchronization}

When recording tactics in interactive mode, it is important to synchronize the
tactic database with the undo/redo actions of the user, both from a theoretical
and practical perspective. In theory, if a user undoes a proof step, this
represents a mistake made by the user, meaning that the recorded information in
the database is also a mistake. In practice, keeping such information will lead
to problems in compilation mode because the database will be smaller due to the
lack of undo/redo actions. Therefore \texttt{search}es that succeeded in
interactive mode may not succeed in compilation mode. Below, we explain how Coq
and Tactician deal with state synchronization.

Internally, Coq ships with a state manager that keeps track of all state
information generated when vernacular commands are executed. This information
includes, for example, definitions, proofs, and custom tactic definitions. This
data is automatically synchronized with the user's interactive movement through
the document, and saved to the binary \texttt{.vo} file during compilation. All
data structures registered with the state manager are expected to be persistent
(in the functional programming sense~\cite{DBLP:journals/jcss/DriscollSST89}).
The copy-on-write semantics of such data structures allow the state manager to
easily keep a history of previous versions and revert to them on demand.

For Tactician, registering data structures with the state manager to ensure
proper synchronization is awkward, because the state manager assumes that
tactics have no side-effects outside of modifications to the proof state. Hence,
any data registered with the state manager is discarded as soon as the current
proof has been finished. Tactician solves this by tricking Coq into thinking
that tactics are side-effecting vernacular commands, convincing it to re-execute
all tactics at \texttt{Qed} time to properly register the side-effects. However,
as a consequence, these tactics will modify the proof state a second time, at a
time when this is not intended. This is a likely source of future bugs for which
a permanent solution is yet to be found.

\section{Tactician as a Machine Learning Platform}
\label{sec:tactician-as-a-machine-learning-platform}
Apart from serving as a tool for end-users, Tactician also functions as a
machine learning platform. A simple OCaml interface to add a new learning model
to Tactician is provided. The learning task of the model is to predict a list of
tactics for a given proof state. When registering a new model, Tactician will
automatically take advantage of it during proof search. Our interface hides
Coq's internal complexities while being as general as possible. We encourage
everyone to implement their favorite learning technique and try to beat the
built-in model. Tactician's performance can easily be benchmarked on the
standard library and other packages. The signature of a machine learning model
is as follows:
\begin{minted}{ocaml}
type sentence = Node of string * sentence list
type proof_state =
{ hypotheses : (id * sentence) list
; goal       : sentence }

type tactic
val tactic_sentence : tactic -> sentence
val local_variables : tactic -> id list
val substitute      : tactic -> id_map -> tactic

module type TacticianModelType = sig
  type t
  val create : unit -> t
  val add : t ->  before:proof_state -> tactic -> after:proof_state -> t
  val predict : t -> proof_state -> (float * tactic) list
end
val register_learner : string -> (module TacticianLearnerType) -> unit
\end{minted}

A \texttt{sentence} is a very general tree data type able to encode all of Coq's
internal syntax trees, such as those of terms and tactics. Node names of syntax
trees are converted into \texttt{string}s. This way, most semantic information
is preserved using a much simpler data type that is suitable for most machine
learning techniques. Proof states are encoded as a list of named hypothesis
sentences and a goal sentence. In this case, sentences represent a Gallina term.
We abstract from some of Coq's proof state complexities such as the shelf and
the unification map.

Tactician represents \texttt{tactic}s as an abstract type that can be inspected
as a sentence using \texttt{tactic\_sentence}. Since the goal of this interface
is to \textit{predict} tactics but not \textit{synthesize} tactics, it is not
possible to modify them (this would seriously complicate the interface). There
is one exception. We provide a way to extract a list of variables that refer to
the local context of a proof. The local variables of a tactic can also be
updated using a simultaneous substitution. Such substitutions will allow for a
limited form of parameter prediction.

We think that local variable prediction is the only kind of parameter prediction
that makes sense in Tactician's context. The only other major classes of
parameters are global lemma names and complete Gallina terms. Predicting
complete terms is known to be very difficult and would unnecessarily complicate
the interface. Predicting names of global lemmas is possible, but does not
appear to be very useful because lemma names are almost always directly
associated with basic tactics like \texttt{apply lem} or \texttt{rewrite lem}.
Predicting parameters for these tactics is counter-productive because their
semantics are mostly dependent on the definition of the lemma. Hence, it is
better to view the incarnations of such tactics with different lemmas as
arguments as completely separate tactics.

Finally, implementing a learning model entails implementing the module type
\texttt{TacticianModelType} and registering it with Tactician. This module
requires an implementation of a database type \texttt{t}. For reasons explained
in \Cref{sec:state-synchronization}, this database needs to be persistent.
Tactician will \texttt{add} \texttt{tactic}s to the database, together with the
\texttt{proof\_state} before and after the tactic was applied. The machine
learning task of the model is to \texttt{predict} a weighted list of tactics
that are likely applicable to a previously unseen proof state.

The current interface only allows for models that support online learning
because database entries are \texttt{add}ed one by one in interactive mode. We
justify this by the user-friendliness requirements from
\Cref{sec:design-decisions}. However, we realize that together with the
persistence requirement, this places considerable limitations on the kind of
learning models that can be employed. In the future, we intent to support a
secondary interface that can be used to create offline models employing batch
learning on large Coq packages in its entirety.

\section{Case Study}
\label{sec:case-studies}
The overall performance of our tactical search on the full Coq Standard Library
is reported in a previous publication~\cite{blaauwbroek2020tactic}, which also
reports performance on various parts of the library. The best-performing version
of our learning model can prove 34.0\% of the library lemmas when using a 40s
time limit. Six different versions together prove 39.3\%. The union with all
CoqHammer methods achieves 56.7\%.\footnote{CoqHammer's eight methods prove
  together 40.8\%, with the best proving 28.8\%.}

Here we show an example of a nontrivial proof found by Tactician. The system was
asked to automatically find the proof of the following lemma from the library
file
\texttt{Structures/GenericMinMax.v},\footnote{\url{https://coq.inria.fr/library/Coq.Structures.GenericMinMax.html}}
where facts about general definitions of \texttt{min} and \texttt{max} are
proved.
\begin{minted}{coq}
Lemma max_min_antimono f :
 Proper (eq==>eq) f -> Proper (le==>flip le) f ->
 forall x y, max (f x) (f y) == f (min x y).
\end{minted} 
Tactician's learning model evaluated the following two lemmas as similar to what
has to be proven:
\begin{minted}{coq}
Lemma min_mono f :
   (Proper (eq ==> eq) f) -> (Proper (le ==> le) f) ->
   forall x y, min (f x) (f y) == f (min x y).
 intros Eqf Lef x y.
 destruct (min_spec x y) as [(H,E)|(H,E)]; rewrite E;
  destruct (min_spec (f x) (f y)) as [(H',E')|(H',E')]; auto.
 - assert (f x <= f y) by (apply Lef; order). order.
 - assert (f y <= f x) by (apply Lef; order). order.
Qed.

Lemma min_max_modular n m p :
   min n (max m (min n p)) == max (min n m) (min n p).
 intros. rewrite <- min_max_distr.
 destruct (min_spec n p) as [(C,E)|(C,E)]; rewrite E; auto with *.
 destruct (max_spec m n) as [(C',E')|(C',E')]; rewrite E'.
 - rewrite 2 min_l; try order. rewrite max_le_iff; right; order.
 - rewrite 2 min_l; try order. rewrite max_le_iff; auto.
Qed.
\end{minted}
The trace through the proof search tree that resulted in a proof is as follows:
\begin{verbatim}
max_min_antimono  .0.0.0.5.5.2.1.0.5.1.5.1        
\end{verbatim}
This trace represents, for every choice point in the search tree, which of
\texttt{suggest}'s ranked suggestion was used to reach the proof. The proof
search went into depth 12 and the first three tactics used in the final proof
are those with the highest score as recommended by the learning model, which
most likely followed the proof of \texttt{min\_mono}. However, after that, it
had to diverge from that proof, using only the sixth-best ranked tactic twice in
a row. This nontrivial search continued for the next seven tactical steps,
combining mostly tactics used in the two lemmas and some other tactics. The
search finally yielded the following proof of \texttt{max\_min\_antimono}.
\begin{minted}{coq}
intros Eqf Lef x y. destruct (min_spec x y) as [(H, E)|(H, E)]. rewrite E.
destruct (max_spec (f x) (f y)) as [(H', E')| (H', E')].
assert (f y <= f x) by (apply Lef; order). order. auto. rewrite E.
destruct (max_spec (f x) (f y)) as [(H', E')| (H', E')]. auto.
assert (f x <= f y) by (apply Lef; order). order.
\end{minted}
Note that the original proof of the lemma is quite similar, but shorter and
without some redundant steps. Redundant steps are known to happen in systems
like Tactician, such as the TacticToe~\cite{DBLP:conf/lpar/GauthierKU17} system
for HOL4~\cite{DBLP:conf/tphol/SlindN08}.
\begin{minted}{coq}
intros Eqf Lef x y. destruct (min_spec x y) as [(H,E)|(H,E)]; rewrite E;
destruct (max_spec (f x) (f y)) as [(H',E')|(H',E')]; auto.
- assert (f y <= f x) by (apply Lef; order). order.
- assert (f x <= f y) by (apply Lef; order). order.
\end{minted}



\section{Related Work}
\label{sec:related-work}
There exist quite  a few machine learning systems for  Coq and other interactive
theorem  provers. The  most significant  distinguishing factor  of Tactician  to
other systems for Coq is its
user-friendliness. 
There are several other systems that  are interesting, but rather challenging to
install and use for end-users. They often depend on
external tools such as machine  learning toolkits and automatic theorem provers.
Some   systems   need   a   long   time  to   train   their   machine   learning
models---preferably on  dedicated hardware. Those  are often not  geared towards
end-users at all but rather towards the Artificial Intelligence community.

Tactician       takes       its       main      inspiration       from       the
TacticToe~\cite{DBLP:conf/lpar/GauthierKU17}              system             for
HOL4~\cite{DBLP:conf/tphol/SlindN08}  which  learns  tactics  expressed  in  the
Standard ML language. Using this knowledge, it can then automatically search for
proofs by  predicting tactics and their  arguments. Our work is  similar both in
doing a  learning-guided tactic search  and by  its complete integration  in the
underlying  proof  assistant  without   relying  on  external  machine  learning
libraries and specialized hardware.

Below is a short list of machine learning systems for the Coq theorem prover.
\begin{description}
\item[ML4PG]  provides   tactic  suggestions  by  clustering   together  various
  statistics               extracted              from               interactive
  proofs~\cite{DBLP:journals/corr/abs-1212-3618}.  It  is  integrated  with  the
  Proof  General~\cite{DBLP:conf/tacas/Aspinall00}  proof  editor  and  requires
  connections to Matlab or Weka. 
\item[SEPIA]  provides  proof  search  using  tactic  predictions  and  is  also
  integrated   with   Proof  General~\cite{DBLP:conf/cade/GransdenWR15}.   Note,
  however, that  its proof search  is only based on  tactic traces and  does not
  make predictions based on the proof state.
\item[Gamepad] is a  framework that integrates with the Coq  codebase and allows
  machine learning  to be performed  in Python~\cite{DBLP:conf/iclr/HuangDSS19}.
  It uses  recurrent neural networks to  make tactic prediction and  to evaluate
  the quality of a  proof state. The system is able to  synthesize proofs in the
  domain of algebraic rewriting. Gamepad is not geared towards end-users.
\item[CoqGym] extracts tactic  and proof state information on a  large scale and
  uses it to  construct a deep learning model capable  of generating full tactic
  scripts~\cite{DBLP:conf/icml/YangD19}.  CoqGym's evaluation  is  using a  time
  limit of 600s, which is impractically high for Coq practitioners. Still, it is
  significantly weaker than CoqHammer. A probable  cause is the slowness of deep
  neural networks which is common to most proving experiments geared towards the
  deep learning community.
\item[Proverbot9001]  is  a proof  search  system  for  Coq  based on  a  neural
  architecture~\cite{DBLP:journals/corr/abs-1907-07794}. The system is evaluated
  on  the verified  CompCert  compiler~\cite{DBLP:journals/cacm/Leroy09}. It  is
  reported that  Proverbot9001's architecture is a  significant improvement over
  CoqGym.
\item[CoqHammer]  is  a  machine  learning  and  proving  tool  in  the  general
  \emph{hammers}~\cite{DBLP:journals/jfrea/BlanchetteKPU16,DBLP:journals/jar/KaliszykU14,DBLP:journals/jar/BlanchetteGKKU16,DBLP:journals/jar/KaliszykU15a}
  category   designed    for   Coq~\cite{DBLP:journals/jar/CzajkaK18}.   Hammers
  capitalize      on     the      capabilities     of      automatic     theorem
  provers~\cite{DBLP:conf/lpar/Schulz13,DBLP:conf/tacas/MouraB08,DBLP:conf/cade/RiazanovV99}
  to assist  ITP's. To  this end,  learning-based premise  selection is  used to
  translate an ITP  problem to the ATP's target language  (typically First Order
  Logic).  A proof  found by  the ATP  can then  be reconstructed  in the  proof
  assistant. CoqHammer is  a maintained system that is  well-integrated into Coq
  and only requires a connection to an ATP. 
  It has  similar performance as  Tactician but proves different  lemmas, making
  these systems complementary~\cite{blaauwbroek2020tactic}.
\end{description}

\section{Further Work and Conclusion}
\label{sec:conclusion}
We have presented Tactician, a seamless and interactive tactic learner and
prover for Coq. The machine learning perspective has been described in a
previous publication~\cite{blaauwbroek2020tactic}. We showed how Tactician is an
easy-to-use tool that can be employed by the user with minimal installation
effort. A clear approach to using the system in large developments has also been
outlined. With its current machine learning capabilities, we expect Tactician to
help the user with its proving efforts significantly. Finally, we presented a
powerful machine learning interface that will allow researchers to bring their
advanced learning models to Coq users while being isolated from Coq's internal
complexities. We expect this to be of considerable utility to both the
artificial intelligence community and Coq users.

There are many future research directions. There is a never-ending quest to
improve the built-in learning model. With better features and stronger (but
still fast) learners such as boosted trees (used in
ENIGMA-NG~\cite{DBLP:conf/cade/ChvalovskyJ0U19})
we hope to push Tactician's performance 
over the standard library towards 50\%. Apart from this, we expect to improve
support for SSreflect and to introduce support for the new Ltac2 language. In
the future, the machine learning interface will be expanded to allow for batch
learning. Additionally, we would like to incorporate the tactic history
(\textit{memory}) of the current lemma into the learning model, similar to
SEPIA. Short-term memory will allow Tactician to modify its suggestions based on
the tactics that were previously executed in the current proof.

\bibliographystyle{plainurl}
\bibliography{bibliography}
\end{document}